\documentclass[conference]{IEEEtran}
\IEEEoverridecommandlockouts
\usepackage{subcaption}
\usepackage{tikz}

\newcommand\copyrighttext{%
  \footnotesize \textcopyright 2025 IEEE. Personal use of this material is permitted.
  Permission from IEEE must be obtained for all other uses, in any current or future
  media, including reprinting/republishing this material for advertising or promotional
  purposes, creating new collective works, for resale or redistribution to servers or
  lists, or reuse of any copyrighted component of this work in other works.}
\newcommand\copyrightnotice{%
\begin{tikzpicture}[remember picture,overlay]
\node[anchor=south,yshift=10pt] at (current page.south) 
  {\fbox{\parbox{\dimexpr\textwidth-\fboxsep-\fboxrule\relax}{\copyrighttext}}};
\end{tikzpicture}%
}
\usepackage{listings}
\usepackage{cite}
\usepackage{amsmath,amssymb,amsfonts}
\usepackage{algorithmic}
\usepackage{graphicx}
\usepackage{textcomp}
\usepackage{makecell}
\usepackage{xcolor}

\usepackage[nolist]{acronym}

\begin{acronym}[MACHU]
  \acro{ai}[AI]{Artificial Intelligence}
  \acro{mdp}[MDP]{Markov Decision Process}
  \acro{soa}[SOA]{Service-oriented Architecture}
  \acro{toh}[ToH]{Towers of Hanoi}
  \acro{rl}[RL]{Reinforcement Learning}
  \acro{fbrl}[FBRL]{Forward-Backward Reinforcement Learning}
  \acro{ncm}[NCM]{Neural Cognitive Modelling}
  \acro{lnn}[LNN]{Logical Neural Networks}
  \acro{llm}[LLM]{Large Language Model}
  \acro{pddl}[PDDL]{Planning Domain Definition Language}
  \acro{fcn}[FCN]{Fully-connected Network}
  \acro{ddqn}[DDQN]{Double Deep Q networks}
  \acro{lif}[LIF]{Leaky Integrate-and-Fire}
  \acro{vsa}[VSA]{Vector Symbolic Architecture}
  \acro{ml}[ML]{Machine Learning}
  \acro{mbrl}[MBRL]{Relational Model-based Reinforcement Learning}
  \acro{ilp}[ILP]{Inductive Logic Programming}
  \acro{strips}[STRIPS]{Stanford Research Institute Problem Solver}
\end{acronym}

\def\BibTeX{{\rm B\kern-.05em{\sc i\kern-.025em b}\kern-.08em
    T\kern-.1667em\lower.7ex\hbox{E}\kern-.125emX}}
\begin{document}

\title{Analysis of AI Techniques for Orchestrating Edge-Cloud Application Migration\\
}

\author{
    \IEEEauthorblockN{
        Sadig Gojayev, Ahmad Anaqreh, Carolina Fortuna
    }
    \IEEEauthorblockA{
        \textit{Department of Communication Systems}, \textit{Jožef Stefan Institute}, SI-1000 Ljubljana, Slovenia\\
        sadiqqocayev10@gmail.com, \{ahmad.anaqreh, carolina.fortuna\}@ijs.si
    }
}

\maketitle
\copyrightnotice
\begin{abstract}
Application migration in edge-cloud system enables high QoS and cost effective service delivery.
However, automatically  orchestrating such migration is typically solved with heuristic approaches. 
Starting from the \ac{mdp}, in this paper, we identify, analyze and compare selected state-of-the-art \ac{ai} planning and \ac{rl} approaches for solving the class of edge-cloud application migration problems that can be modeled as  \ac{toh} problems. We introduce a new classification based on state space definition and analyze the compared models also through this lense. The aim is to understand available techniques capable of orchestrating such application migration in emerging computing continuum environments. 
\end{abstract}


\section{Introduction} \label{sec:intro}
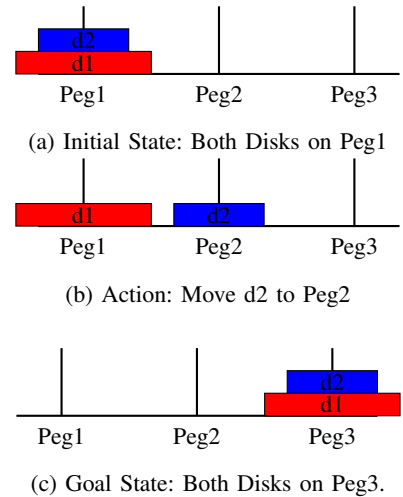
\begin{figure}[t]
    \centering
    \begin{subfigure}[b]{0.45\textwidth}
        \centering
        \begin{tikzpicture}[scale=0.6] 
            \draw[thick] (0,0) -- (0,1.5); 
            \draw[thick] (3,0) -- (3,1.5); 
            \draw[thick] (6,0) -- (6,1.5); 

            \draw[thick] (-1,0) -- (7,0);

            \draw[fill=red] (0-1.5,0) rectangle (0+1.5,0.5); 
            \draw[fill=blue] (0-1.0,0.5) rectangle (0+1.0,1.0); 

            \node[scale=0.9] at (0,-0.5) {Peg1}; 
            \node[scale=0.9] at (3,-0.5) {Peg2};
            \node[scale=0.9] at (6,-0.5) {Peg3};

            \node[scale=0.9] at (0,0.75) {d2};
            \node[scale=0.9] at (0,0.25) {d1};
        \end{tikzpicture}
        \caption{Initial State: Both Disks on Peg1}
    \end{subfigure}
    \hspace{0.05\textwidth}
    \begin{subfigure}[b]{0.45\textwidth}
        \centering
        \begin{tikzpicture}[scale=0.6] 
            \draw[thick] (0,0) -- (0,1.5); 
            \draw[thick] (3,0) -- (3,1.5); 
            \draw[thick] (6,0) -- (6,1.5); 

            \draw[thick] (-1,0) -- (7,0);

            \draw[fill=red] (0-1.5,0) rectangle (0+1.5,0.5); 
            \node[scale=0.9] at (0,0.25) {d1};

            \draw[fill=blue] (3-1.0,0) rectangle (3+1.0,0.5); 
            \node[scale=0.9] at (3,0.25) {d2};

            \node[scale=0.9] at (0,-0.5) {Peg1};
            \node[scale=0.9] at (3,-0.5) {Peg2};
            \node[scale=0.9] at (6,-0.5) {Peg3};
        \end{tikzpicture}
        \caption{Action: Move d2 to Peg2}
    \end{subfigure}
    
    \vspace{0.5cm}
    
    \begin{subfigure}[b]{0.45\textwidth}
        \centering
        \begin{tikzpicture}[scale=0.6] 
            \draw[thick] (0,0) -- (0,1.5); 
            \draw[thick] (3,0) -- (3,1.5); 
            \draw[thick] (6,0) -- (6,1.5); 

            \draw[thick] (-1,0) -- (7,0);

            \draw[fill=red] (6-1.5,0) rectangle (6+1.5,0.5); 
            \draw[fill=blue] (6-1.0,0.5) rectangle (6+1.0,1.0); 

            \node[scale=0.9] at (0,-0.5) {Peg1};
            \node[scale=0.9] at (3,-0.5) {Peg2};
            \node[scale=0.9] at (6,-0.5) {Peg3};

            \node[scale=0.9] at (6,0.75) {d2};
            \node[scale=0.9] at (6,0.25) {d1};
        \end{tikzpicture}
        \caption{Goal State: Both Disks on Peg3.}
    \end{subfigure}
    \caption{Initial, Action, and Goal States for Two Disks with Pegs.}
    \label{fig:TOH}
\end{figure}

Application migration  involves transferring computational tasks between cloud to close/far edge computing  infrastructures \cite{ray2020proactive}. This process is important for reducing local processing loads, conserving energy, and executing intensive operations that exceed the device's capabilities, thereby enhancing system performance and scalability in distributed environments. Consider a smart city scenario where a multi-component application update must be migrated from a central cloud to a local edge device through an intermediate regional edge server. Each component (or microservice) of the update has dependencies—for example, core services must be updated before supplementary ones. Additionally, due to limited bandwidth or processing capacity, as is often the case in resource-constrained environments,   sequential, dependency-aware migration planning across distributed computing nodes is needed\cite{he2023taxonomy}. Such a scenario can be modeled as a \ac{toh} problem. \ac{toh} involves discs and pegs, where all the discs are initially placed on the first peg, and the goal is to move all the discs to the last peg. The key limitation is that only a smaller disc can be placed on top of a larger disc, as illustrated in Figure~\ref{fig:TOH}.

Assuming the application follows a \ac{soa}, each microservice (or update package) represents a "disk" depicted in Figure \ref{fig:TOH}. Larger disks (core components) have higher priority and strict dependency requirements compared to smaller disks (supplementary components). The three computing layers—the central cloud, the regional edge server, and the local edge device—act as the three pegs. The goal is to move all components from the central cloud (source peg) to the local edge device (target peg), using the regional server (auxiliary peg) as an intermediary. Just like in the \ac{toh} where a larger disk cannot be placed on top of a smaller one, the update process requires that a component with higher dependency (core service) must not be delayed or “blocked” by a later (or smaller) update. The migration must occur in an order that respects these dependencies.

Automatically orchestrating application migration in cloud-edge environments in an efficient way is a challenging task \cite{bisicchia2024continuous}. To date, most of the works consider heuristics for solving such problems \cite{li2024energy}. However, various AI methods, from the field of AI planning and \ac{rl} could also be suitable candidates. More specifically, for the considered example that solves the \ac{toh} problem, several automatic methods have been considered in the \ac{ai} planning \cite{pallagani2023plansformer,stewart2011neural} and \ac{rl} \cite{suttonreinforcement,edwardsforward} literature. However, to date it is unclear how the approaches from the two communities can be compared to each other \cite{lee2022ai}. Perhaps the common denominator is the fact that both communities aim to find automated ways of solving problems that can be modeled by a \ac{mdp}. 

Starting from the \ac{mdp}, in this paper, we identify, analyze and compare selected state-of-the-art \ac{ai} planning and \ac{rl} approaches for solving the class of edge-cloud application migration problems that can be modeled as  \ac{toh} problems. The aim is to understand available techniques capable of orchestrating such application migration in computing continuum environments. The contributions of this paper are:

\begin{itemize}
    \item We introduce a new categorization as \textit{state space definition}, complementing the recently introduced \textit{access level} categorization.
    \item We identify and summarize LLM, RL, search and solve, spiking neural network and logical neural network based solutions for solving the problem.
    \item We analyze these models through the \ac{mdp} perspective and categorize based on access levels, approach to state definition  and compare in terms of time cost and validity, offering insights into their scalability and efficiency.    
\end{itemize}

This paper is organized as follows: Section~\ref{sec:mdp_intro} discusses the \ac{mdp}, provides an overview of access levels and introduces a new classification based on the approach to state definition. Section~\ref{sec:methods} summarizes and differentiates various automated models for solving \ac{toh} problem. Section~\ref{sec:anlyz_methods} presents an analysis of the methods, framed through the elements of the \ac{mdp} model, access levels and state definition. Finally, Section~\ref{sec:conclusion} concludes the paper.

\section{\ac{mdp}} \label{sec:mdp_intro}
Both \ac{ai} planning and \ac{rl} focus on an agent's ability to achieve a goal. This goal-seeking process can be formulated as a \textit{\ac{mdp}}, which provides a general framework for decision-making. An \ac{mdp} is defined as \cite{chen2024ai}:

\begin{equation}
    M = \langle S, A, T, R, G \rangle,
\end{equation}

States ($S$) represent the environment's configurations, and actions ($A$) define possible agent moves. The transition function ($T$) describes the probability of reaching a new state $s'$ after taking action $a$ in state $s$. The reward function ($R$) assigns values to transitions, guiding the agent toward desirable outcomes, while goal states ($G$) mark the completion of the process.

A policy $\pi: S \to A$ defines the agent’s actions, and the value function $V_{\pi}(s)$, defined by the Bellman equation \cite{barron1989bellman}, measures the expected cumulative reward of policy $\pi$.

\begin{equation}
    \scalebox{0.8}{$
    V_{\pi}(s) = \begin{cases}
        0, & \text{if } s \in G, \\
        \sum_{s' \in S} T(s, \pi(s), s') \left( R(s, \pi(s), s') + V_{\pi}(s') \right), & \text{otherwise}.
    \end{cases}
    $}
\end{equation}

An optimal policy $\pi^*$ maximizes the value function across all states:

\begin{equation}
    V_{\pi^*}(s) = \max_{\pi} V_{\pi}(s), \quad \forall s \in S.
\end{equation}

The \ac{toh} problem from Figure \ref{fig:TOH} can be formulated as a \ac{mdp} with states representing all possible configurations of disks on three pegs. Each action consists of moving the top disk from one peg to another, provided it follows the game’s rules (i.e., no larger disk on a smaller one). Transitions occur deterministically when an action is taken, leading to a new valid state. A typical reward function could assign -1 per move to encourage solving the problem in the fewest steps. The goal state is reached when all disks are moved from the source peg to the destination peg in the correct order, achieving optimality in minimal moves \( 2^n - 1 \) for \( n \) disks.

\subsection{Access Levels}
A recent study \cite{chen2024ai} raised an important question regarding the level of access an agent has to the environment, proposing a classification of access levels within a hierarchical framework. Based on this, five levels of \ac{mdp} access are defined:

     \paragraph{Continuing} The agent has no prior knowledge of $T$ or $R$ and experiences the environment in a single continuous sequence. Learning occurs purely through trial and error.
    \paragraph{Episodic} Similar to the continuing case, but the environment resets after reaching a terminal state.
    \paragraph{Generative} The agent has access to a simulator that provides samples from $T$, allowing for data collection without full knowledge of the environment.
    \paragraph{Analytic} The agent has complete knowledge of $T$ and $R$, enabling direct computation of optimal policies without trial-and-error learning.
    \paragraph{Structured} Beyond full knowledge of $T$ and $R$, the agent can exploit structured representations of the problem to improve decision-making efficiency.

\subsection{State Space Definition} 
\label{sec:observation}

After analyzing the agent's access levels, we identified another critical aspect: the initialization states. To systematically categorize how models handle state constraints during problem-solving, we introduce a \textbf{State Space Definition}, classifying methods into three distinct categories:
    \paragraph{Predefined States (PS)} Training uses a precomputed dataset of valid states, so no exploration of invalid states is needed.
    
    \paragraph{Rule-based State Generation (RSG)} The state space rules are defined before training, ensuring only valid states are generated.
    
    \paragraph{Constraint-based  State Filtering (CSF)} The model encounters both valid and invalid states during learning, dynamically refining its decision-making to filter out invalid states based on constraint enforcement simultaneously.

\section{Automated Methods for Solving \ac{toh}} \label{sec:methods}

In this section we identify and  analyze different methods for solving the Towers of Hanoi problem: the LLM-based Plansformer, \ac{fbrl}, search and solve  Neurosolver, spiking neural network based \ac{ncm} and \ac{lnn}. 

\subsection{Plansformer}

Plansformer \cite{pallagani2023plansformer} was designed to generate symbolic plans through fine tuning an \ac{llm} model CodeT5 \cite{wang2021codet5}. A plan, as defined by this approach, is expressed in the \ac{pddl} \cite{fox2003pddl2}, which contains [GOAL], [INIT], [ACTION], [PRE], [EFFECT] to represent the goal state, initial state, possible actions with their associated preconditions and effects these actions cause in the environment, respectively, with an example for ToH depicted in Figure ~\ref{fig:PDDL_syntax}.

\begin{figure}[ht]
    \centering
    \begin{minipage}{0.9\columnwidth}
    \begin{lstlisting}
<GOAL> on d1 peg2, clear d1, on d2 peg1, clear d2, clear peg3
<INIT> smaller peg1 d1, smaller peg1 d2, smaller peg2 d1, smaller peg2 d2,
smaller peg3 d1, smaller peg3 d2, smaller d2 d1, on d1 d2, clear d1,
on d2 peg3, clear peg1, clear peg2
<ACTION> move
<PRE> smaller to disc, on disc from, clear disc, clear to
<EFFECT> clear from, on disc to, not on disc from, not clear to
    \end{lstlisting}
    \end{minipage}
    
    \caption{\ac{pddl} Description with Syntax Highlighting.}\footnotemark
    
    \label{fig:PDDL_syntax}
    
\end{figure}

\footnotetext{https://github.com/VishalPallagani/LLMsforPlanningLab-AAAI24/tree/main/Part\%204/data/test\_samples}

By relying on \ac{llm}s, the planning is approached as a classification task over the vocabulary it was tuned on. Based on prompting with a problem description, the GOAL, initial state, INIT, possible set of actions, ACTION, preconditions, PRE, and their effects, EFFECT, they generate a sequence of tokens forming a plan.

\subsubsection{Training Data} 
The \ac{toh} dataset consists of 18,000 synthetically generated plans with varying configurations, with an average plan length of 12 moves. For training, 5-fold cross-validation is applied with an 80\%-20\% split, and a vocabulary size of 32,005 after tokenization.

\subsubsection{\ac{ml} Model} 
CodeT5, a masked language model with an encoder-decoder architecture, is used for plan generation. Pre-trained on code generation tasks to effectively leverage syntactic information, its weights are then updated to align with plan generation. CodeT5 is fine-tuned using a training dataset of 14,400 samples.

\subsubsection{Evaluation Metrics and Results}
The model generates a plan for a new problem instance, which must be evaluated in terms of sentence-level structure and validity, where validity refers to the plan’s ability to solve the problem. ROUGE-L and BLEU were used to assess model performance, with ROUGE-L achieving 0.97 and BLEU 0.95, indicating high sentence-level similarity. For planner performance, VAL \cite{howey2004val} evaluates validity and optimality, with Plansformer achieving 84.97\% validity and 82.58\% optimality in an average time of 5 seconds for the \ac{toh}.

\subsection{\ac{fbrl}}
\begin{figure}[htbp]
    \centering
    \includegraphics[width=0.31\textwidth]{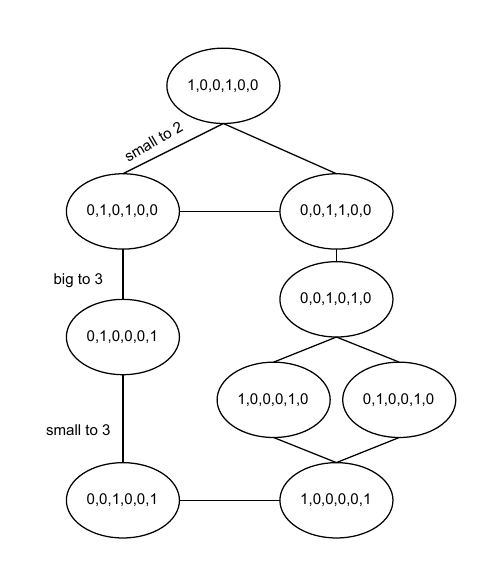} 
    \caption{Possible States of \ac{toh} with 2 Disks.}
    \label{fig:toh_states}
\end{figure}

 \ac{fbrl} extends typicall RL  by enabling the agent to reason backward in time, improving learning efficiency in sparse reward environments \cite{edwardsforward}. In the \ac{toh} problem, states represent the positions of discs on pegs. These states are represented as bit strings, as illustrated in Figure~\ref{fig:toh_states}.

Each group of 3 bits in a string represents the location of a disc. For 2 discs, the state is represented by 6 bits. For example, [1,0,0,1,0,0] denotes the initial state, and [0,0,1,0,0,1] represents the goal state, as illustrated in Figure~\ref{fig:TOH}. Actions involve moving a disc from one peg to another while maintaining the problem constraints, and rewards are assigned based on progress toward the goal state. The transition model in \ac{fbrl} is based on a backward dynamics model, which predicts the previous state by reversing actions, allowing the agent to simulate transitions both forward and backward. By iteratively exploring from the start state and the goal state, \ac{fbrl} spreads reward signals more effectively, accelerating the learning process compared to standard model-free approaches.

\subsubsection{Training data}
A replay buffer is a mechanism used in \ac{rl} to store and manage the agent's experiences (state, action, reward, next state, and done flag) and the forward step is to train the agent using experiences from the replay buffer, according to some learning paradigm, here \ac{ddqn} is used \cite{van2015deep}. 

\ac{fbrl} utilizes both imagined and real experiences to learn values. A forward step uses samples of real experiences originating from the start state to update Q-values, and a backward step uses imagined states that are asynchronously predicted in reverse from known goal states. Real experiences are transitions directly collected from agent-environment interactions, while imagined experiences are synthetic transitions generated by a learned backward dynamics model to simulate how the agent might have reached the goal.

\subsubsection{\ac{ml} Model}
\ac{fbrl} model architecture is a \ac{fcn} which is also known as multi-layer perceptron \cite{li2018fully} with 100 outputs followed by RELU, followed by another \ac{fcn} with 9 outputs, representing the distribution over each bit.

\subsubsection{Evaluation Metrics and Results}
\ac{fbrl} is compared against a standard \ac{ddqn} baseline \cite{van2015deep} by comparing the average reward. The average reward is calculated based on mean cumulative reward obtained by the agent over a set number of epochs. The findings show an advantage for using \ac{fbrl} as the goal gets further away and increasing the number of discs, \ac{fbrl} outperforms \ac{ddqn}. It degraded for 3 discs where \ac{fbrl} has around -0.5 average reward while \ac{ddqn} has an average 0.8 in 30000 training steps, which may be due to overfitting.

\subsection{Neurosolver Learning}

Neurosolver is a neuromorphic planner that models problem-solving as a network of interconnected nodes, where each node represents a state and connections define valid transitions \cite{bieszczad2015neurosolver}. It constructs a state-space memory by learning relationships between states during training, enabling it to find an optimal path to the goal. In the \ac{toh} problem, states are represented as strings, where each peg's disk configuration is encoded (e.g., for 2 disks, [12,00,00] as the initial state and [00,00,12] as the goal). Neurosolver learns valid transitions and, using this memory, dynamically determines a solution by searching backward from the goal state to the current state.

\subsubsection{Training data}
In the random training phase, Neurosolver autonomously explores the problem space by generating random but valid state transitions like [023, 000, 001] → [003, 002, 001].  It systematically moves between states while ensuring that each move adheres to the constraints of the problem (e.g., no larger disk on a smaller one in \ac{toh}). Each encountered state is stored as a node and transitions between states are recorded as connections between nodes. Over multiple training cycles, Neurosolver constructs a state-space network, gradually mapping out all possible moves

\subsubsection{\ac{ml} Model}
\begin{figure}[htbp]
    \centering
    \includegraphics[width=0.5\textwidth]{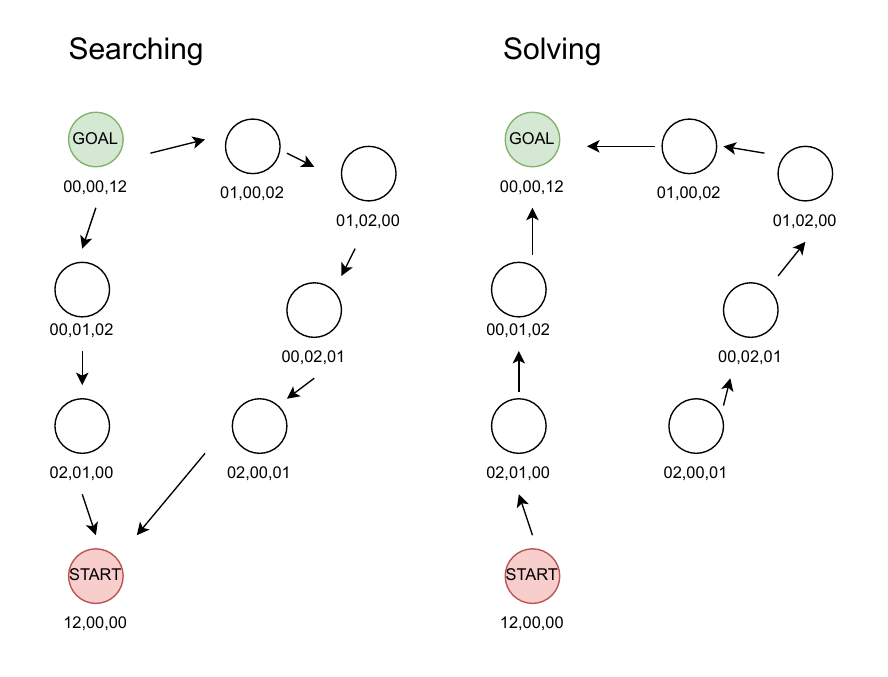} 
    \caption{Neurosolver Searching and Solving Phase for \ac{toh} Problem with 2 Disks.}
    \label{fig:toh_neurosolver}
\end{figure}

The nodes consist of two parts, known as the upper and lower divisions, which work together to perform the primary tasks of the Neurosolver: searching and solving. In the searching phase, it starts from the goal state and propagates activity backward through its network of nodes to find the shortest path to the current state, ensuring an optimal solution. In the solving phase, it activates nodes step by step, following the planned route from the current state to the goal. The upper division guides the sequence by selecting the next move, while the lower division executes it, systematically solving the puzzle as illustrated in Figure \ref{fig:toh_neurosolver}.

\subsubsection{Evaluation Metrics and Results}
Neurosolver was evaluated based on its ability to complete tasks within a time cost relative to the size of the problem. Firstly, it implemented a search part with 1,000,000 training samples and showed that up to 10 discs finding possible states lasts around 125 seconds. It took 1,000,000 steps vs. 10,000 which were sufficient to create a model for the four disc version showing optimal path length 15.

\subsection{\ac{ncm}}

\ac{ncm} \cite{stewart2011neural} is a brain-inspired computer simulation that solves the \ac{toh} puzzle using spiking neurons using \ac{lif} \cite{teeter2018generalized}. The model follows neural processes to make decisions, store memory by \ac{vsa} \cite{schlegel2022comparison}, and plan moves. The model tracks disk positions with three neural groups: WHAT (disk being moved), WHERE (target location), and ATTEND (currently focused disk for checking obstacles). States and actions are represented as high-dimensional vectors, while rewards are encoded through synaptic connections. The model operates by selecting actions based on the neural representations of these states, using the basal ganglia, thalamus, and cortex for planning and action selection.

\subsubsection{Training data}
The design and parameters of the model are derived from biological constraints and algorithmic implementation of the \ac{toh} task, it relies on predefined neural structures and rules based on biological principles. The model's "training" is essentially the configuration of synaptic weights and neural connections that represent cognitive processes like memory storage, goal recall, and action selection.
\subsubsection{\ac{ml} Model}
The model uses 150,000 \ac{lif} neurons to simulate brain activity with discrete spikes, representing aspects like disk positions, movement rules, and goal states for biologically plausible decision-making. Action selection follows the basal ganglia-thalamus-cortex loop, and the model operates with 19 predefined rules encoded in neural activation patterns and synaptic weights, dictating disk movement, obstacle management, and subgoal generation. A key feature is its goal memory, utilizing \ac{vsa} to prevent redundant calculations, enabling efficient problem-solving by remembering disk destinations even after multiple intermediate steps.

\subsubsection{Evaluation Metrics and Results}
The evaluation metrics for the neural model focus on reaction time and efficiency. The model solves a segment of the \ac{toh} problem in approximately 0.9 seconds, with 0.2 seconds for goal recall, 0.2 seconds for disk attention, and 0.5 seconds for executing two moves. These metrics assess the model's ability to mimic human-like reaction times and perform decision-making within biologically realistic time frames.

\subsection{\ac{lnn}}

The paper \cite{agravante2023learning} focuses on learning \ac{strips} operators using \ac{lnn} for \ac{mbrl} using \ac{ilp} \cite{muggleton1991inductive}. \ac{strips} \cite{fikes1971strips} is a formalism used in \ac{ai} planning to represent actions in terms of preconditions and effects, allowing an agent to plan sequences of actions efficiently. \ac{lnn} learn STRIPS operators by identifying logical rules from \ac{rl} interactions, where each action is broken down into four components: \textbf{preconditions} ($\alpha$), \textbf{negative preconditions} ($\beta$), \textbf{positive effects} ($\gamma$), and \textbf{negative effects} ($\delta$). The \ac{toh} problem is represented using relational logic, where states are expressed with predicates as shown in Table \ref{tab:STRIPS_Hanoi}.

\begin{table}[h]
    \caption{STRIPS Representation of the Move Operator in Towers of Hanoi.}

    \centering
    \begin{tabular}{|p{2.5cm}|p{2.5cm}|} 
        \hline
        \textbf{Component} & \textbf{Logical Representation} \\
        \hline
        \textbf{Preconditions} ($\alpha$) & \texttt{clear(disc) $\wedge$ on(disc, from)  $\wedge$ clear(to)} \\
        \hline
        \textbf{Negative Preconditions} ($\beta$) & \texttt{$\neg$ on(from, disc)} \\
        \hline
        \textbf{Positive Effects} ($\gamma$) & \texttt{on(disc, to) $\wedge$ clear(from)} \\
        \hline
        \textbf{Negative Effects} ($\delta$) & \texttt{$\neg$ on(disc, from)} \\
        \hline
    \end{tabular}
    \label{tab:STRIPS_Hanoi}
\end{table}

Rewards are tied to goal achievement, and actions are planned to maximize them. State transitions are determined by applying learned \ac{strips} operators through \ac{rl}. This allows for planning using symbolic representations while benefiting from neural networks' ability to handle complex data.

\subsubsection{Training Data}
The dataset is collected from \ac{rl} interactions, where an agent explores the environment by taking actions and observing state transitions. Each interaction generates tuples $\langle s_t, a_t, s_{t+1}, r_t \rangle$, where states and actions are represented in relational logic. The data is then structured into positive and negative examples, with preconditions extracted from the initial state and effects derived from state differences.

\subsubsection{ML Model}
The model uses \ac{ilp} within \ac{lnn} to learn STRIPS operators by optimizing logical rules from \ac{rl} data. \ac{ilp} infers preconditions and effects by processing relational state transitions. Each \ac{strips} component is represented as a logical rule \( h \leftarrow b_1 \land b_2 \land \dots \land b_n \), where \( h \) is the conclusion and \( b_1, b_2, \dots, b_n \) are conditions to be satisfied. The dataset includes positive examples (\( P \)) for valid transitions and negative examples (\( N \)) for invalid ones. Unlike classical \ac{ilp}, \ac{lnn}s use real-valued logic, assigning weights to predicates (e.g., \(\text{on(disk1, peg1)}\)) rather than treating them as strictly true or false. These weights are optimized via gradient-based learning, enabling refinement of logical rules in noisy environments.

\subsubsection{Evaluation Metrics and Results}
The model is evaluated based on the validity of success in solving the problem. The \ac{lnn}-learned model achieves 100\% success rates in solving problems in \ac{toh}.

\section{Analysis of the Methods} \label{sec:anlyz_methods}
In this section we map the \ac{ai} based methods discussed in Section~\ref{sec:methods} to the \ac{mdp}, access level and state space categories from Section~\ref{sec:mdp_intro} .

\subsection{From the \ac{mdp} Model Perspective}
As shown in Table~\ref{tab:comparison_mdp}, the \ac{mdp} elements—State, Action, Transition, Reward, and Goal—are represented in the columns, with each row corresponding to methods.

\paragraph*{Plansformer}
From Table~\ref{tab:comparison_mdp} it can be seen that in the Plansformer the \textbf{state space (S)} consists of symbolic representations encoded in  \ac{pddl}. It defines the \texttt{[GOAL]}, \texttt{[INIT]}, and the current environment configuration. For example, in \ac{toh}, a state may be represented as \texttt{on d1 peg2, clear d1, on d2 peg1, clear d2, clear peg3}. The \textbf{action space (A)} includes symbolic planning operations such as \texttt{move}, where each action has \texttt{[PRE]} (preconditions) and \texttt{[EFFECT]} (state changes) explicitly defined. The \textbf{transition function (T)} is learned using a fine-tuned \ac{llm}, specifically CodeT5, which classifies and generates valid plans based on the input problem definition. The \textbf{reward function (R)} is implicitly structured by the model’s training objective, where plans are evaluated based on syntactic correctness and execution validity, aligning with expert-generated plans. The \textbf{goal state (G)} is defined as the state where all disks are placed on the target peg, matching the symbolic \texttt{[GOAL]} specification in \ac{pddl}.

\begin{table*}[ht]
\caption{Comparison of \ac{toh} models in terms of their State, Action, Transition, Reward, and Goal Representations.}
\centering
\scriptsize
\setlength{\tabcolsep}{1.5pt} 
\renewcommand{\arraystretch}{1.7} 
\resizebox{\textwidth}{!}{ 
\begin{tabular}{|l|p{1.5cm}|p{2cm}|p{4.5cm}|p{2cm}|p{2.5cm}|}
    \hline
    \textbf{Model} & \textbf{State (S)} & \textbf{Action (A)} & \textbf{Transition (T)} & \textbf{Reward (R)} & \textbf{Goal (G)} \\
    \hline
    \textbf{Plansformer} & Symbolic rules (\ac{pddl}) & Symbolic moves with conditions & $T(s, a) \Rightarrow s'$ via CodeT5 plan generation. & Based on correct, executable plans & All disks on the target peg (\ac{pddl} goal conditions). \\
    \hline
    \textbf{\ac{fbrl}} & Bit string (disk positions) & Valid disk moves (bit changes) &$Q(S_t, A_t) \approx \mathbb{E} [R + \gamma \max Q(S_{t+1}, A)]$
     & +1 valid, -1 invalid, +10 for goal & Bit string for disks on Peg C (binary pattern). \\
    \hline
    \textbf{Neurosolver} & Tuples (disk positions) & Legal disk moves between pegs & $s' = f(s, a)$ learned via neural exploration. & Rewards progress, penalizes errors & Tuple of disks on the target peg (iterative search). \\
    \hline
    \textbf{\ac{ncm}} & Neural patterns (disk positions) & 19 predefined disk moves & $s' = W \times s + b$ (spiking neuron update). & Implicit—reinforces valid moves & Stable neural pattern for goal state (spiking activity). \\
    \hline
    \textbf{\ac{lnn}} & Logic + neural embeddings & Logical moves (with rules) & $s' = \arg\max f_\theta(s, a)$ (weighted logic update). & Rewards valid moves, penalizes errors & Logical formula for goal satisfaction. \\
    \hline
\end{tabular}
}

\label{tab:comparison_mdp}
\end{table*}

\paragraph*{\ac{fbrl}}
In \ac{fbrl}, the \textbf{state space (S)} is represented as a bit string of length \(3n\), where each group of 3 bits encodes the peg position of a disk. For example, for 2 disks, the initial state is \([1,0,0,1,0,0]\) (both disks on Peg A), and the goal state is \([0,0,1,0,0,1]\) (both disks on Peg C). The \textbf{action space (A)} consists of valid disk moves, represented as modifications to the corresponding \textbf{3-bit segment} in the bit string while ensuring legal transitions. The \textbf{transition function (T)} is approximated using a \textbf{\ac{ddqn}}, where transitions are learned from a \textbf{replay buffer} and follow the Bellman equation:  
\begin{equation} 
Q(S_t, A_t) \approx \mathbb{E} [R + \gamma \max Q(S_{t+1}, A)]
\end{equation}

The \textbf{reward function (R)} assigns \( +1 \) for valid moves, \( -1 \) for illegal actions (e.g., placing a larger disk on a smaller one), and a \textbf{terminal reward of \( +10 \)} upon reaching the goal. The \textbf{goal state (G)} is the bit-string representation where all disks are correctly transferred to \textbf{Peg C}, marking the successful completion of the problem.

\paragraph*{Neurosolver}

In Neurosolver, the \textbf{state space (S)} is represented by the positions of disks across the three pegs, captured symbolically. For example, a state with 2 disks on Peg A may be represented as \((1,1)\). The \textbf{action space (A)} consists of legal disk movements, defined by transferring the topmost disk between pegs while obeying constraints. The \textbf{transition function (T)} is learned by exploring the state space and generating a self-created dataset of valid transitions. The \textbf{reward function (R)} rewards progress toward the goal state, penalizing illegal moves or redundant transitions, with the highest reward for reaching the goal configuration. The \textbf{goal state (G)} is achieved when all disks are moved to the target peg. Neurosolver focuses on autonomous dataset generation and adaptive learning, allowing iterative policy refinement through internal exploration.

\paragraph*{\ac{ncm}}
In \ac{ncm}, the \textbf{state space (S)} is encoded using the activity of 150,000 Leaky \ac{lif} neurons, simulating peg and disk configurations. Each disk's position is represented by neural activation patterns within the basal ganglia-thalamus-cortex loop, reflecting the current state. For example, a state with both disks on Peg A is encoded by specific neural firing patterns. The \textbf{action space (A)} consists of 19 predefined movement rules, implemented through neural synaptic weight adjustments. The \textbf{transition function (T)} is modeled by dynamic neural activity, with neuron firing propagating changes across layers to represent valid state transitions, using \ac{vsa} for memory encoding. The \textbf{reward function (R)} is embedded in neural dynamics, where successful transitions strengthen synaptic weights, and invalid moves are penalized. The \textbf{goal state (G)} is achieved when a stable neural firing pattern representing the target configuration is reached, completing the decision-making process.
\paragraph*{\ac{lnn}}
In \ac{lnn}, the \textbf{state space (S)} combines symbolic logic and neural embeddings, encoding disk positions using logical predicates (e.g., \texttt{on(d1, peg1)}, \texttt{clear(peg3)}), maintaining interpretability while allowing continuous approximations. The \textbf{action space (A)} consists of logical operators corresponding to valid disk movements, defined by rules (e.g., \texttt{move(d1, peg1, peg3)}) that must satisfy preconditions. The \textbf{transition function (T)} is modeled by differentiable logical inference, where neural networks approximate truth values and apply reasoning to predict the next state based on the current state and action. The \textbf{reward function (R)} gives positive rewards for legal actions that move the system closer to the goal state and penalizes invalid moves, ensuring logical consistency. The \textbf{goal state (G)} is defined by a logical formula where all disks are on the target peg with no rule violations, enabling \ac{lnn} to balance logical soundness with adaptive neural optimization.

\subsection{Access Level, Evaluation Metrics and State Space}
After reviewing the \ac{mdp} components of each method, we analyze their performance across various evaluation metrics, examining how they align with the access and state space definition categories from Sections~\ref{sec:mdp_intro} and~\ref{sec:observation}. Table~\ref{tab:comparison} summarizes the access level, time cost, steps (epochs), validity, and State Space Definition categories in rows for each \ac{ai} model listed in the columns.

\begin{table}[ht]
\caption{Model Comparison.}

\centering
\scriptsize
\begin{tabular}{|c|c|c|c|c|c|}
\hline
\textbf{} & \textbf{\ac{fbrl}} & \textbf{Plansformer} & \textbf{Neurosolver} & \textbf{\ac{lnn}} & \textbf{\ac{ncm}} \\
\hline
\textbf{Access Level} & Gen. & Str. & Gen. & Analytic & Str. \\
\hline
\textbf{Time} & 300 s & 0.05 s & 125.79 s & N/A & 0.9 s \\
\hline
\textbf{Steps (Epochs)} & 30,000 & 10 & 1,000,000 & N/A & N/A \\
\hline
\textbf{Validity} & 100\% & 85\% & 100\% & 100\% & 100\% \\
\hline
\textbf{State Space} & RSG & PS & CSF & CSF & RSG \\
\hline
\end{tabular}
\label{tab:comparison}
\end{table}

\paragraph{\textbf{Access}}
As we saw in Section~\ref{sec:mdp_intro}, access levels depend on a model's approach to transition dynamics and rewards. \ac{fbrl}, for instance, operates under the \textbf{Generative} access level, as it generates transitions and updates its Q-values using both real and imaginary experiences. In contrast, Plansformer falls under the  \textbf{Structured} access level, as its decision-making is guided by a carefully curated dataset, which provides a predefined set of states and transitions. Similarly, Neurosolver also belongs to the  \textbf{Generative} access level, but unlike \ac{fbrl}, it dynamically refines its decision-making process during training, generating its own dataset in the process. \ac{lnn}, on the other hand, are classified under the  \textbf{Analytic} access level, as they have full access to transition dynamics and rewards after training, enabling them to directly compute optimal policies. Finally, \ac{ncm} operates under the  \textbf{Structured} access level, drawing on predefined rules and structured access to transition dynamics and rewards to guide its decision-making process.

\paragraph{\textbf{Time}}
The second key aspect in Table~\ref{tab:comparison} is the comparison of models based on time cost. \ac{fbrl} has a higher time cost, requiring 0.01 seconds per step, leading to a total of 300 seconds for 30,000 epochs. In contrast, Plansformer is highly efficient, generating optimal solutions in 0.05 seconds on average for 3 disks (end-to-end training time is not considered for evaluating plan generation). Neurosolver, though slower, takes 125.79 seconds for 2 disks and 126.49 seconds for 3 disks due to its dynamic learning process. \ac{lnn} excel in solution correctness but lack specific time metrics. Finally, \ac{ncm} performs quickly, solving a segment of the problem in about 0.9 seconds.
\paragraph{\textbf{Steps (Epochs)}}
Another key metric from Table~\ref{tab:comparison} is the comparison of models based on the number of steps (epochs). \ac{fbrl} requires 30,000 epochs to perform well, but its performance struggles as the problem complexity increases. In contrast, Plansformer achieves optimal solutions with only around 10 effective training epochs, thanks to its predefined dataset. Neurosolver, due to its dynamic learning process, requires significantly more steps, using 1,000,000 training samples to refine its decision-making. While \ac{lnn}s and \ac{ncm} don’t provide specific epoch counts, both demonstrate efficient learning.
\paragraph{\textbf{Validity}}
A critical comparison in Table~\ref{tab:comparison} is the models' validity. From \ac{fbrl} to \ac{ncm}, achieve 100\% validity in their solutions. However, Plansformer falls short in this regard, achieving slightly lower validity. This is due to its reliance on LLM-based plan generation, which, while efficient, can occasionally lead to errors or suboptimal solutions due to the inherent limitations of language models in capturing structure and logic.

\paragraph{\textbf{State Space Definition}}
As proposed in Section~\ref{sec:observation}, State Space Definition categories depend on a model's predefined constraints and exploration strategy. Starting with \ac{fbrl}, it follows the \textbf{Rule-based State Generation (RSG)} approach, defining state space rules before training to ensure valid states. In contrast, Plansformer uses the \textbf{Predefined States (PS)} approach, relying on a curated dataset of valid states, however, in the generation process it is able to produce also invalid states. Moving to Neurosolver and \ac{lnn}s, both follow the \textbf{Constraint-based State Filtering (CSF)} approach, dynamically filtering invalid states as they learn, with \ac{lnn}s gaining the advantage of full access to transition dynamics and rewards. Finally, \ac{ncm} also employs the \textbf{RSG} approach, using predefined rules and structured state spaces.

\section{Conclusions} \label{sec:conclusion}

In this work we aimed to better understand available automated techniques for orchestrating application migration in edge-cloud system to enable high QoS and cost effective service delivery. We identified and summarized LLM, RL, search and solve, spiking neural network and logical neural network based solutions for solving a class of application migration problems that can be modeled as Towers of Hanoi. We analyze these solutions through the \ac{mdp} perspective and categorized them  based on existing access levels taxonomy, proposed state space definition taxonomy and analyze them  in terms of time cost and validity. This work is a prerequisite for developing new automated orchestration functionality for application migration.

\section*{Acknowledgment}
This work was supported by the Slovenian Research Agency (P2-0016) and the European Commission NANCY project (No. 101096456).

\bibliographystyle{IEEEtran}
\bibliography{references}

\begin{thebibliography}{10}
\providecommand{\url}[1]{#1}
\csname url@samestyle\endcsname
\providecommand{\newblock}{\relax}
\providecommand{\bibinfo}[2]{#2}
\providecommand{\BIBentrySTDinterwordspacing}{\spaceskip=0pt\relax}
\providecommand{\BIBentryALTinterwordstretchfactor}{4}
\providecommand{\BIBentryALTinterwordspacing}{\spaceskip=\fontdimen2\font plus
\BIBentryALTinterwordstretchfactor\fontdimen3\font minus
  \fontdimen4\font\relax}
\providecommand{\BIBforeignlanguage}[2]{{%
\expandafter\ifx\csname l@#1\endcsname\relax
\typeout{** WARNING: IEEEtran.bst: No hyphenation pattern has been}%
\typeout{** loaded for the language `#1'. Using the pattern for}%
\typeout{** the default language instead.}%
\else
\language=\csname l@#1\endcsname
\fi
#2}}
\providecommand{\BIBdecl}{\relax}
\BIBdecl

\bibitem{ray2020proactive}
K.~Ray, A.~Banerjee, and N.~C. Narendra, ``Proactive microservice placement and
  migration for mobile edge computing,'' in \emph{2020 IEEE/ACM Symposium on
  Edge Computing (SEC)}.\hskip 1em plus 0.5em minus 0.4em\relax IEEE, 2020, pp.
  28--41.

\bibitem{he2023taxonomy}
T.~He and R.~Buyya, ``A taxonomy of live migration management in cloud
  computing,'' \emph{ACM Computing Surveys}, vol.~56, no.~3, pp. 1--33, 2023.

\bibitem{bisicchia2024continuous}
G.~Bisicchia, S.~Forti, E.~Pimentel, and A.~Brogi, ``Continuous qos-compliant
  orchestration in the cloud-edge continuum,'' \emph{Software: Practice and
  Experience}, vol.~54, no.~11, pp. 2191--2213, 2024.

\bibitem{li2024energy}
J.~Li, D.~Zhao, Z.~Shi, L.~Meng, W.~Gaaloul, and Z.~Zhou, ``Energy-efficient
  online service migration in edge networks,'' \emph{IEEE Internet of Things
  Journal}, 2024.

\bibitem{pallagani2023plansformer}
V.~Pallagani, B.~Muppasani, K.~Murugesan, F.~Rossi, L.~Horesh, B.~Srivastava,
  F.~Fabiano, and A.~Loreggia, ``Plansformer: Generating symbolic plans using
  transformers,'' in \emph{NeurIPS 2023 Workshop on Generalization in
  Planning}.

\bibitem{stewart2011neural}
T.~Stewart and C.~Eliasmith, ``Neural cognitive modelling: A biologically
  constrained spiking neuron model of the tower of hanoi task,'' in
  \emph{Proceedings of the Annual Meeting of the Cognitive Science Society},
  vol.~33, no.~33, 2011.

\bibitem{suttonreinforcement}
R.~S. Sutton, ``Reinforcement learning: An introduction,'' \emph{A Bradford
  Book}, 2018.

\bibitem{edwardsforward}
A.~D. Edwards, L.~M. Downs, and J.~C. Davidson, ``Forward-backward
  reinforcement learning.''

\bibitem{lee2022ai}
J.~Lee, M.~Katz, D.~J. Agravante, M.~Liu, T.~Klinger, M.~Campbell, S.~Sohrabi,
  and G.~Tesauro, ``Ai planning annotation for sample efficient reinforcement
  learning.''

\bibitem{chen2024ai}
D.~Z. Chen, P.~Verma, S.~Srivastava, M.~Katz, and S.~Thi{\'e}baux, ``Ai
  planning: A primer and survey (preliminary report),'' \emph{arXiv preprint
  arXiv:2412.05528}, 2024.

\bibitem{barron1989bellman}
E.~Barron, ``The bellman equation for minimizing the maximum cost.''

\bibitem{wang2021codet5}
Y.~Wang, W.~Wang, S.~Joty, and S.~C. Hoi, ``Codet5: Identifier-aware unified
  pre-trained encoder-decoder models for code understanding and generation,''
  in \emph{Proceedings of the 2021 Conference on Empirical Methods in Natural
  Language Processing}.\hskip 1em plus 0.5em minus 0.4em\relax Association for
  Computational Linguistics, 2021.

\bibitem{fox2003pddl2}
M.~Fox and D.~Long, ``Pddl2. 1: An extension to pddl for expressing temporal
  planning domains,'' \emph{Journal of artificial intelligence research},
  vol.~20, pp. 61--124, 2003.

\bibitem{howey2004val}
R.~Howey, D.~Long, and M.~Fox, ``Val: Automatic plan validation, continuous
  effects and mixed initiative planning using pddl,'' in \emph{Proceedings of
  the 16th IEEE International Conference on Tools with Artificial
  Intelligence}, 2004, pp. 294--301.

\bibitem{van2015deep}
H.~v. Hasselt, A.~Guez, and D.~Silver, ``Deep reinforcement learning with
  double q-learning,'' in \emph{Proceedings of the Thirtieth AAAI Conference on
  Artificial Intelligence}, 2016, pp. 2094--2100.

\bibitem{li2018fully}
J.~Li, B.~Li, J.~Xu, R.~Xiong, and W.~Gao, ``Fully connected network-based
  intra prediction for image coding,'' \emph{IEEE Transactions on Image
  Processing}, vol.~27, no.~7, pp. 3236--3247, 2018.

\bibitem{bieszczad2015neurosolver}
A.~Bieszczad and S.~Kuchar, ``Neurosolver learning to solve towers of hanoi
  puzzles,'' in \emph{2015 7th International Joint Conference on Computational
  Intelligence (IJCCI)}, vol.~3.\hskip 1em plus 0.5em minus 0.4em\relax IEEE,
  2015, pp. 28--38.

\bibitem{teeter2018generalized}
C.~Teeter, R.~Iyer, V.~Menon, N.~Gouwens, D.~Feng, J.~Berg, A.~Szafer, N.~Cain,
  H.~Zeng, M.~Hawrylycz \emph{et~al.}, ``Generalized leaky integrate-and-fire
  models classify multiple neuron types,'' \emph{Nature communications},
  vol.~9, no.~1, p. 709, 2018.

\bibitem{schlegel2022comparison}
K.~Schlegel, P.~Neubert, and P.~Protzel, ``A comparison of vector symbolic
  architectures,'' \emph{Artificial Intelligence Review}, vol.~55, no.~6, pp.
  4523--4555, 2022.

\bibitem{agravante2023learning}
D.~J. Agravante, D.~Kimura, and M.~Tatsubori, ``Learning neuro-symbolic world
  models with logical neural networks,'' in \emph{PRL Workshop Series Bridging
  the Gap Between AI Planning and Reinforcement Learning}, 2023.

\bibitem{muggleton1991inductive}
S.~Muggleton, ``Inductive logic programming,'' \emph{New generation computing},
  vol.~8, pp. 295--318, 1991.

\bibitem{fikes1971strips}
R.~E. Fikes and N.~J. Nilsson, ``Strips: A new approach to the application of
  theorem proving to problem solving,'' \emph{Artificial intelligence}, vol.~2,
  no. 3-4, pp. 189--208, 1971.

\end{thebibliography}
\end{document}